\documentclass[runningheads]{llncs}

\usepackage{eccv}

\usepackage{eccvabbrv}

\usepackage{graphicx}
\usepackage{booktabs}
\usepackage{amsmath}
\usepackage{amssymb}
\usepackage{hhline}
\usepackage{tikz}
\usepackage{listings}
\usepackage{makecell}
\usepackage{multirow}
\usepackage{float}
\usepackage{xcolor}
\usepackage{caption}
\usepackage{siunitx}
\usepackage{enumitem}
\usepackage{mathrsfs}
\usepackage{tabularx}
\usepackage{authblk}

\usepackage[accsupp]{axessibility}  % Improves PDF readability for those with disabilities.

\newif\ifcomments
\commentstrue
\usepackage[pagebackref,breaklinks,colorlinks]{hyperref}
\title{FBINeRF: Feature-Based Integrated Recurrent Network
for Pinhole and Fisheye Neural Radiance Fields} 
\titlerunning{FBINeRF}
\author{Yifan Wu\inst{1} \and Tianyi Cheng\inst{1} \and Peixu Xin\inst{1} \and Janusz Konrad\inst{1}} 
\authorrunning{YF.Wu, ~et al.} 
\institute{Department of Electrical and Computer Engineering, Boston University \\
\email{\{wuyifan1, chengty, brucexin, jkonrad\}@bu.edu}}

\begin{document}

\maketitle
\begin{abstract}
Previous studies aiming to optimize and bundle-adjust camera poses using Neural Radiance Fields (NeRFs), such as BARF and DBARF, have demonstrated impressive capabilities in 3D scene reconstruction. However, these approaches have been designed for pinhole-camera pose optimization and do not perform well under radial image distortions such as those in fisheye cameras. Furthermore, inaccurate depth initialization in DBARF results in erroneous geometric information affecting the overall convergence and quality of results. In this paper, we propose adaptive GRUs with a flexible bundle-adjustment method adapted to radial distortions and incorporate feature-based recurrent neural networks to generate continuous novel views from fisheye datasets. Other NeRF methods for fisheye images, such as SCNeRF and OMNI-NeRF, use projected ray distance loss for distorted pose refinement, causing severe artifacts, long rendering time, and are difficult to use in downstream tasks, where the dense voxel representation generated by a NeRF method needs to be converted into a mesh representation. We also address depth initialization issues by adding MiDaS-based depth priors for pinhole images. Through extensive experiments, we demonstrate the generalization capacity of FBINeRF and show high-fidelity results for both pinhole-camera and fisheye-camera NeRFs.
\end{abstract}

\section{Introduction}
% \label{sec:intro}
Neural Radiance Field (NeRF) \cite{mildenhall2020nerf} is a state-of-the-art method that reconstructs a 3D scene from 2D images and can create highly-realistic and detailed renderings from new perspectives. NeRF employs a neural network encoder to minimize the difference between rendered views and actual dataset images. Additionally, it uses a multilayer perceptron (MLP) to create a scene representation by evaluating a 5D implicit function that estimates scene density and radiance.

\begin{figure}[h]
\centering
\includegraphics[scale=0.45]{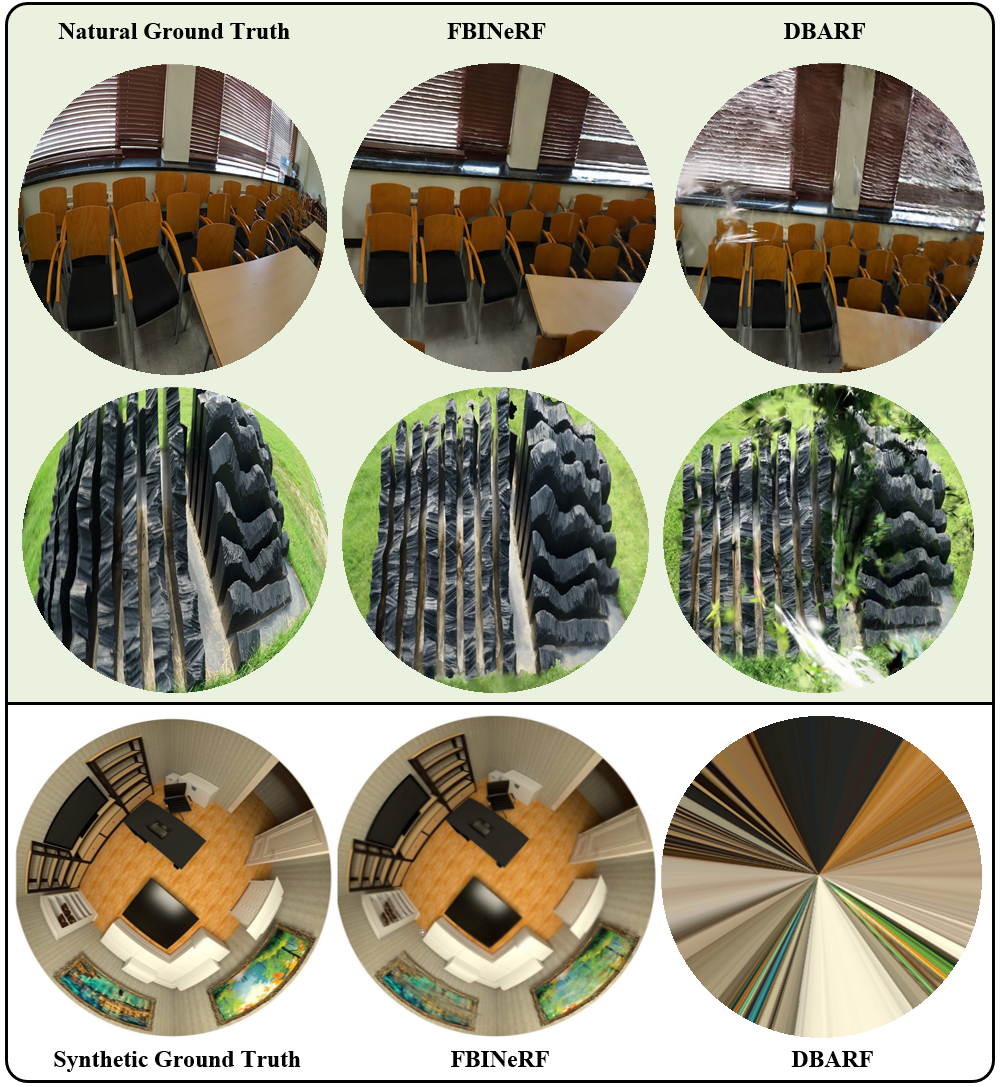} 
\caption{\textbf{FBINeRF and DBARF reconstructions for natural and synthetic fisheye data}. Two examples from FisheyeNeRF dataset \cite{jeong2021selfcalibrating} demonstrate that even a very small radial distortions degrade performance of DBARF but not of FBINeRF. \textbf{Top:} Chairs image - DBARF reconstruction (no lens-distortion model) shows severe distortions on some chairs and window blinds. \textbf{Middle:} Rock image - obvious distortions especially at periphery. \textbf{Bottom}: Synthetic image \cite{eichenseer2022data} - cannot be reconstructed by DBARF.}
%\vspace{-15pt} 
% \caption{An overview of relative and metric depth combined with flexible bundle adjusting to render novel synthetic views. RGB images are input into the MiDaS depth estimation framework\cite{bhat2023zoedepth} to obtain both relative and metric depth information. Initial pose estimates and sparse 3D points are derived using COLMAP. Subsequently, the refined camera poses and depth maps from flexible bundle adjusting are utilized to project 3D points onto 2D images. The culmination of these steps leads to the rendering of the final NeRF scenes.}
\label{fig 1}
\centering
\end{figure}

NeRF relies on relatively precise initial camera pose information. However, in practice the acquisition of camera poses is expensive. In previous studies, camera poses have been usually acquired by means of Structure-from-Motion (SfM) methods \cite{zhu2017parallel,lindenberger2021pixelperfect,7780814}. In SfM and its variants, camera poses are optimized under a keypoint-metric reprojection error in a process referred to as bundle adjustment \cite{Triggs1999BundleA}. SfM may encounter difficulties, \textit{e.g.,} in scenes lacking texture or containing self-similar elements, and can often be time-consuming, requiring days to complete large-scale scenes. As a result, a significant challenge for NeRF is its strong dependence on precise camera poses in order to produce high-quality renderings. Recent studies, such as BARF \cite{lin2021barf}, have proposed to jointly train camera-pose recovery with NeRF. Inspired by non-smoothness optimization in high-dimensional functions, BARF employs a coarse-to-fine approach initially eliminating high-frequency components and then gradually reactivating them once low-frequency components have reached stability. The resulting smoother objective function eases training although the optimizer may still get trapped in a local optimum. During training, camera poses are fine-tuned using a photometric loss rather than relying on keypoint-metric cost as in SfM. Despite these innovations, BARF still relies on pre-computed camera poses.

More recently, generalizable NeRF methods \cite{wang2021ibrnet,chen2021mvsnerf,chen2023dbarf} have drawn much attention. Most of them, unlike BARF, directly extract features for optimization instead of positional encoding. DBARF \cite{chen2023dbarf} jointly bundle-adjusts camera poses by leveraging a smooth feature cost map (from Feature Pyramid Network) as an implicit cost function in a self-supervised manner. The contribution of DBARF is that it combines camera poses with generalizable NeRF methods and thus can be applied to NeRF not optimized per-scene, without accurate initial camera poses. However, DBARF has limited ability to tackle geometrically-distorted images (Fig. \ref{fig 1}). NeRF methods to-date involve camera-pose optimization \textbf{using global optimization with geometric constraints leading to slow convergence of distortion parameters and severe artifacts}. Moreover, they \textbf{initialize depth maps {\it via} a mapping from features to depth values} \cite{chen2023dbarf,gu2023dro}, thus potentially resulting in reduced accuracy and difficulties with convergence when dealing with distorted camera poses, eventually reducing their generalization ability.

We propose \textit{FBINeRF} to overcome the issues mentioned above. We use IBRNet as a baseline for NeRF rendering and a deep recurrent network for relative pose refinement. Pinhole and fisheye NeRF rendering are separated. In the fisheye camera pipeline (our main contribution), we utilize adaptive GRUs to process image features extracted from DenseNet\cite{huang2018densely} and update the pose with incremental hidden states. Then, a flexible bundle adjustment block is adapted to radial image distortions when jointly training relative camera poses with feature cost map in the recurrent neural network. This enables faster convergence during training compared to traditional fisheye NeRF methods and generates dense and continuous synthetic views from fisheye images (see videos in supplementary materials). For the pinhole camera pipeline, inspired by ZoeDepth \cite{bhat2023zoedepth}, we add MiDaS-like depth priors (self-supervised) and metric-bins module (supervised) to initialize depth for better convergence and adaptation to unseen data. \textit{FBINeRF} switches to different models according to the input image data via designated types.

In summary, we present a novel integrated generalizable network that generates dense voxel fisheye novel synthetic views that can be converted into mesh representations for downstream tasks in Blender, Unity, or Unreal Engine 5 where simulations, such as for autonomous vehicles, may require meshes or point clouds. We also introduce depth priors to fix DBARF's convergence issues caused by random depth initialization in the pinhole camera model and enhance the pinhole model's generalization ability to produce depth priors from unseen datasets.

\section{Related Work}
%\textbf{Depth Estimation}
\subsubsection{Depth Estimation} Conventional approaches \cite{Henry2010RGBDMU,7299195,LONGUETHIGGINS198761} use stereo vision, where depth is computed from feature correspondence between multiple images taken from different viewpoints. These methods are limited in their ability to handle occlusions and textureless regions. In recent years, deep learning has revolutionized depth estimation by employing large-scale datasets. For example, in \cite{godard2017unsupervised} a self-supervised learning framework was introduced that exploits left-right consistency in stereo pairs to train a depth-estimation model without requiring ground-truth depth annotations. In \cite{zhang2018deep}, a deep-learning approach was proposed that combines RGB and sparse depth data to produce dense high-resolution depth maps.
%a network architecture was proposed to fuse information from multiple modalities.
%More recently, depth estimation approaches have been divided into two mainstream, either relative depth estimation(RDE) \cite{Lee_2019_CVPR, Mertan_2022, ranftl2020robust}, or metric depth estimation(MDE) \cite{Farooq_Bhat_2021, bhat2022localbins, li2022binsformer}. Inspired by \cite{bhat2023zoedepth}, FBINeRF leverage both.
The most recent depth-estimation research can be grouped into two categories: relative depth estimation \cite{Lee_2019_CVPR, Mertan_2022, ranftl2020robust} and metric depth estimation \cite{Farooq_Bhat_2021, bhat2022localbins, li2022binsformer}. Inspired by results reported in \cite{bhat2023zoedepth}, we leverage both types of methods in FBINeRF.
% FBINeRF can leverages both relative and metric depth estimation to equip the NeRF model with better generalization when using depth as training guidance. It then contributes to the accuracy of the rendering process. 

\subsubsection{Neural Radiance Fields} NeRF methods developed to-date have used diverse approaches to improve rendering quality or speed. Some methods focus on mitigating aliasing effects \cite{barron2021mipnerf,barron2022mipnerf,barron2023zipnerf}. Other methods aim to reduce both training and rendering time \cite{wang2023f2nerf, chen2022tensorf, M_ller_2022, sun2022direct, yu2021plenoxels}. Pixel-NeRF \cite{yu2021pixelnerf} is a benchmark method that first generalized NeRFs to unseen views. It utilizes projection and interpolation to construct a feature volume containing features extracted from images. A subsequent MLP network renders RGB and density values. IBRNet \cite{wang2021ibrnet} combines image features on a per-point basis using a weighting scheme inspired by PointNet \cite{qi2017pointnet}. It extends that approach by introducing a ray transformer \cite{vaswani2023attention} to estimate density and an additional MLP to predict pixel colors. MVSNeRF \cite{chen2021mvsnerf} builds a 3D feature-cost volume based on $N$ depth hypotheses, and employs a 3D CNN to reconstruct a neural-voxel volume followed by an MLP to predict pixel colors and volume density. Another category of NeRF methods  applies {\it pose refinement} \cite{yenchen2021inerf,wang2022nerf,meng2021gnerf,zhang2023vmrf,chng2022garf,chen2023dregnerf}. NeRF- - \cite{wang2022nerf} jointly optimizes a NeRF network and camera-pose embeddings, and achieves comparable accuracy to NeRF methods that require posed images. SiNeRF \cite{xia2022sinerf} avois the sub-optimality issue in NeRF- - by using a mixed-region sampling strategy. VMRF \cite{zhang2023vmrf} allows to learn NeRF without known camera poses. An unbalanced optimal transport is incorporated to model the relative transformation between the actual and rendered image; the predicted relative poses are utilized to update camera poses, thus enabling a more refined NeRF training.

\subsubsection{NeRF with Distortion Rectification} OMNI-NERF \cite{shen2022omninerf} extends the capabilities of traditional NeRF models to omnidirectional imagery, allowing it to handle scenes captured with omnidirectional cameras. SCNeRF \cite{jeong2021selfcalibrating} addresses the challenge of fisheye-lens distortions up to mild-distortion levels. However, the method attempts to capture geometric information from basic NeRF model to assure photometric consistency, which limits the potential for better optimization and rectification of distortions. In contrast, our method, FBINeRF, takes advantage of generalized NeRF models for pose refinement to help tackle distortions.

\begin{figure}[h]
\centering
\includegraphics[scale=0.21]{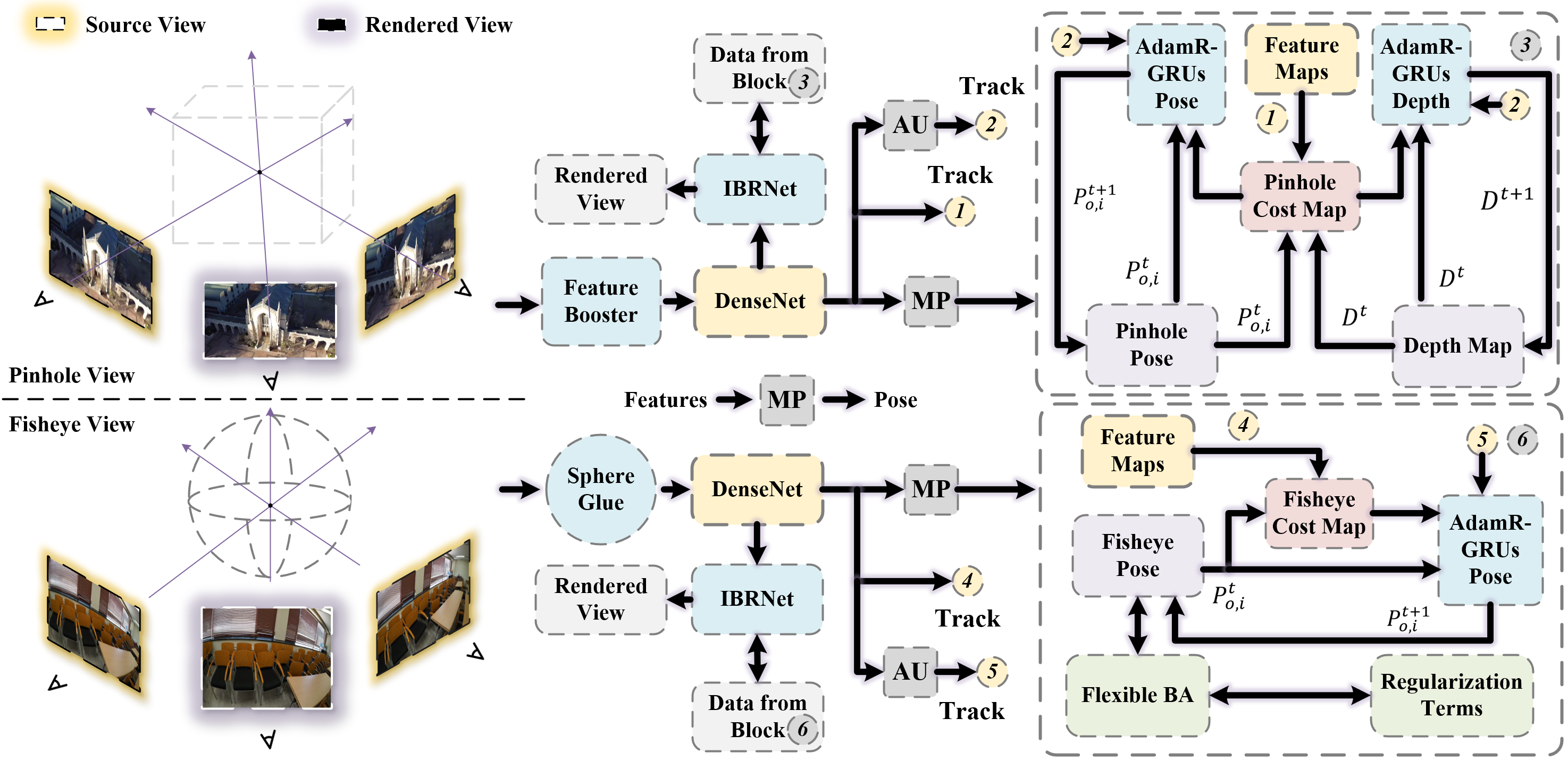} 
\caption{\textbf{Network architecture of FBINeRF.} 1) The input can be either pinhole or fisheye views. Neighbor views are selected by keypoint matching: FeatureBooster \cite{wang2023featurebooster} in pinhole views and SphereGlue \cite{inproceedings} in fisheye views. 2) Image features are extracted by DenseNet-like \cite{huang2018densely} backbone and sent to subsequent blocks. 3) Depth priors are utilized for depth map initialization in pinhole model and kept updating; Metric-bins module enhances predicted depth priors through fine-tuning with supervised learning, please refer to \cite{bhat2023zoedepth}. 4) Pose/depth, image features, contextual features, cost map, and context vectors are sent to AdamR-GRU \cite{PAL2023110457} optimizer to update poses and depth values. For fisheye views, depth update is replaced by flexible bundle adjustment with a lens-distorsion model to update camera poses. 5) IBRNet is used as the baseline to predict color and density values, jointly training with both pose-optimization blocks.}
%\vspace{-10pt} 
\label{fig 2}
\centering
\end{figure}

\section{Methods}

\subsection{Selection of Neighbor Poses}
Existing generalized NeRF methods aggregate features from nearby views by means of the nearest top-$k$ views. For planar keypoint matching in pinhole images, we use FeatureBooster \cite{wang2023featurebooster}, a lightweight network that enhances feature descriptors, followed by nearest-neighbor (NN) search. Traditional spherical keypoint matching fails to leverage global keypoint neighbor information leading to erroneous or insufficient matches. We use SphereGlue \cite{inproceedings} to handle geometric distortions in fisheye images. We use the epipolar constraint \cite{szeliski2022computer} as a "sanity check" to validate our network, only for debugging purpose. Ground truth poses are not used during training.

\subsection{Pinhole-Pose Refinement}
Given a pinhole image $\mathbf{\mathcal{I}}_o$ and its $N$ nearby views $\{\mathcal{I}_{i}\}^{N}_{i=1}$, image features are extracted using DenseNet and are fed into a mapping block (MP in Fig.~\ref{fig 2}) to obtain the initial relative camera pose $\mathcal{P}^{init}_{o,i}$ from image $o$ to image $i$, while MiDaS-like depth priors are predicted for depth initialization. Then, image pose/depth, cost map, contextual features and context vectors (\textit{via} attention unit, AU in Fig. \ref{fig 2}) are sent to AdamR-GRUs-like \cite{PAL2023110457} optimizer for either pose or depth refinement, in order to minimize photometric loss projected into feature space \cite{tang2018banet}. As the feature cost map that measures the distance between aligned feature maps, we use the average of multiple-cost values inspired from DBARF \cite{chen2023dbarf}:
\begin{equation}
\small
     C(x) = \frac{1}{N}\sum^N_{i=1} ||\mathcal{F}_i(\Pi(\mathcal{T}_i \circ \Pi^{-1}(x, D(x))))-\mathcal{F}_o(x)||_2
\end{equation}

\noindent where $\Pi(\cdot)$ is a mapping of 3D point to the image plane, and its inverse $\Pi^{-1}(\cdot)$ maps an image pixel $x$, given its depth $D(x)$, to a point in 3D space. $\mathcal{T}_i$ transforms 3D points between coordinate spaces of two images, e.g., $\mathcal{I}_o$ and $\mathcal{I}_i$. The iterative steps to minimize the feature cost map $C$ are \cite{gu2023dro}:
\begin{equation}
\begin{aligned}
    D^{t+1} &= D^t + \mathcal{M}(D^t,C^t, \mathcal{F}_{o}^C,{c}^t_o) \\
    \mathcal{P}_{o,i}^{t+1} &= \mathcal{P}_{o,i}^t + \mathcal{M}(\mathcal{P}_{o,i}^t,C^t, \mathcal{F}_{i}^C,{c}^t_i)
\end{aligned}
\end{equation}
where $\mathcal{M}(\cdot)$ is a mapping in AdamR-GRUs \cite{PAL2023110457} optimizer involving either the depth $D$ or relative camera pose $\mathcal{P}_{o,i}$, cost map $C$, contextual feature $\mathcal{F}^C$, and context vector $c$. The context vector $c$ is derived from the attention unit (see below) and the contextual feature $\mathcal{F}^C$ is obtained {\it via} similar feature-extraction backbone with a different mapping.

\subsubsection{DenseNet and Feature Matrix} DenseNet-121 \cite{huang2018densely} is used as a backbone to extract features with shared feature weight from different images, denoted $\mathcal{F}_o$ for image $\mathcal{I}_o$ and $\{\mathcal{F}_i\}_{i=1}^{N}$ for nearby views. 
The extracted features form a high-dimensional vector, where each element originates from all prior layers in DenseNet. More specifically, the output of the $l^{th}$ layer is: $Q_l = \mathcal{T}_q(q_0,q_1,\ldots,q_{l-1})$,
where $\{q_0,q_1,\ldots,q_{l-1}\}$ denotes the concatenation of features extracted from previous layers and $\mathcal{T}_q$ denotes a non-linear transformation. To reduce the computational complexity, transition layers are introduced as a bottleneck \cite{PAL2023110457}. We map the features into a variable-length feature matrix $\mathcal{A}$ that is later reshaped through attention units to match the dimensions of the input to AdamR-GRU. For images $\mathcal{I}_o$, $\mathcal{I}_i$, respective feature matrices are $\mathcal{A}_o \in \mathbb{R}^{H_o \times W_o}$ and $\mathcal{A}_i \in \mathbb{R}^{H_i \times W_i}$. Each extracted feature $\mathcal{F}_i$ contains contextual feature $\mathcal{F}^C_i$ and feature volume $\mathcal{F}^V_i \in \mathbb{R}^{H_i \times W_i \times d}$, containing $d$ feature matrices $\mathcal{A}_i$ for image $\mathcal{I}_i$.

\subsubsection{Attention Units and AdamR-GRUs} The inputs to attention units consist of each image's feature matrix $\mathcal{A}$, the current attention sum from previous time step $t-1$, and the previous hidden state $h^{t-1}$. The output is a context vector ${c}^t$. We wrap this vector, either the depth value $D$ or pose $\mathcal{P}$, contextual feature $\mathcal{F}^C$ and the feature cost map $\mathcal{C}$ into a tensor denoted $m^t$ at time step $t$. The current pose $\mathcal{P}_{o,i}^t$ and depth $D^t$ are then predicted from hidden state $h^{t}$. Also, in AdamR-GRUs we replace the linear operation with convolution for processing temporal sequences of images and add an adaptive parameter in basic GRUs \cite{gu2023dro} as follows: 
\begin{equation}\label{eqn:gru}
\begin{aligned}
    \hat{v}^{t+1}, \hat{h}^{t+1}, \hat{m}^{t+1} &= \text{AR-G}(h^t, v^t, m^t) \\
    v^{t+1}, h^{t+1}, m^{t+1} &= \text{AR-G}(c^t, \hat{h}^{t+1}, \hat{v}^{t+1}, \hat{m}^{t+1}) 
\end{aligned}
\end{equation}
% \yf{They are intermediate variable used in second equation, I forgot to add hat previously}
where $v^t$ denotes momentum auxiliary state at time $t$, and $c^t$ denotes the current context vector from the attention unit. Finally, either the depth maps or camera poses are iteratively updated using the hidden state $h^t$.

\subsection{Fisheye Pose Refinement}
\label{sec DM}
We introduced FBINeRF for pinhole-camera pose optimization using a deep recurrent neural network. Now, we extend our method to fisheye-pose refinement.
\subsubsection{Distortion Modeling} The key to recover the true ray direction of pixel coordinates on the image plane regardless of distortion is to model the projection of the fisheye lens onto the image plane. The radial distance $r_d$ from the image center to a distorted image point can be calculated as follows  \cite{eichenseer2022data}:

\begin{equation}
    r_d = 2f \cdot \mathrm{sin(\frac{\theta}{2})}
\end{equation}
%
% \yf{It is a standard equisolid mapping (there are 4 types of mappings for distorted images, each of them has different properties, here I select equisolid mapping)} 
where $\theta$ is the incoming angle measured from the lens axis and $f$ is the focal length. To account for different types of camera lenses, we consider an approximation of the incoming ray direction $\theta$ by an odd-order polynomial \cite{szeliski2022computer}:
\begin{equation}
    \theta = \theta_{d} + \sum_{i=1}^3 k_{i} \cdot \theta_{d}^{2i+1}, \quad
    \theta_{d} = \mathrm{arctan}(\frac{r_d}{f})
\end{equation}
%
% \yf{This formula is inconsistent with the one above. Not sure I got this, but $\theta$ is incoming angle between object and optical center, the $\theta_d$ means refraction angle, maybe a figure will show this much better} 
%
\noindent where $k_1,k_2,k_3$ are coefficients modeling fisheye-lens distortions. Given pixel $\mathbf{P}$ with coordinates ($u,v$) on the image plane, we can express the radial distance $r_d$ as follows:
\begin{equation}
    r_d = \sqrt{(u-c_x)^2+(v-c_y)^2}
\end{equation}
%
% \jlk{Why is there $f$ in this formula?}
where ($c_x,c_y$) are the principle-point coordinates. The actual direction of the ray for pixel $\mathbf{P}$, as a vector, is: $\mathbf{d}_c = [\mathrm{sin(\theta)}\cdot(u-c_x),\mathrm{sin(\theta)}\cdot(v-c_y),\mathrm{cos(\theta)}]^T$. Applying rotation $\mathbf R$ and translation $\mathbf t$, ray direction $\mathbf{d}_c$ in world coordinates is expressed as follows:
\begin{equation}
    {\mathbf d} = [{\mathbf R}^{-1}|{\mathbf t}]\; {\mathbf d}_c
\end{equation}
%
% \jlk{Why an inverse rotation?}\yf{To obtain ray direction in world coordinates that is used in NeRF ray calculation. Since d is connected with feature and camera parameters.}
where the rotation matrix $\mathbf{R}$ can be parameterized by an axis-angle representation expressed through parameters $\mathbf{\Phi}$, as explained in \cite{shen2022omninerf}. The unknown camera parameters to estimate are:
\begin{equation}
    \pi_i = (f_x,f_y,c_x,c_y,\mathbf{\Phi}_i,\mathbf{t}_i,k_{1},k_{2},k_{3})
\label{eq pi}
\end{equation}
where $f_x$ and $f_y$ are focal-length to pixel-size ratios along the $x$ and $y$ axes, respectively, and $c_x$ and $c_y$ are coordinates of the principal point with respect to the top-left corner of the image. Intrinsic parameters are known and the poses (in $\pi_i$) are randomly initialized. We jointly optimize camera parameters $\pi_i$, and parameters $\mathbf{\Theta}$ of the IBRNet to handle geometric distortions by minimizing the photometric cost in feature space: 

% The calculated coordinate and direction of a sampled ray are fed into the NeRF model and the rendered color value of the corresponding pixel is derived by the volumetric equation in NeRF \cite{mildenhall2020nerf}. 

%
\begin{equation}
   C(x) = \frac{1}{N}\sum^N_{i=1} ||\mathcal{F}_i(\Pi(\mathcal{T}_i \circ \Pi^{-1}(x,\pi_i))-\mathcal{F}_o(x)||_2
\label{eq fba}
\end{equation}
where $\Pi^{-1}(x,\pi_i)$ represents the inverse mapping of pixel coordinates $x$, given $\pi_i$, to 3D space. Subsequently, each 3D point undergoes transformation $\mathcal{T}_i$ based on camera parameters $\pi_i$. Combining the warped features generated by the IBRNet baseline with the refined fisheye pose from FBINeRF, we can produce continuous and densely synthesized fisheye views. 

\subsubsection{Discussion} The primary distinction between the pinhole camera pipeline and the fisheye camera pipeline lies in the distortion model (exclusive to fisheye cameras) and depth priors (exclusive to pinhole cameras). Both pipelines utilize the same feature extraction method and similar adaptive GRUs for pose refinement. Additionally, they leverage warped features from IBRNet to render novel views.

\subsubsection{Flexible Bundle Adjustment} We utilized flexible bundle adjustment in our fisheye pipeline within feature-based deep recurrent neural network to update fisheye poses. However, due to the finer pose updates required by fisheye cameras, failure to control the step size of updates can lead to the model getting trapped in local optima and failing to converge. Thus, we propose a regularization term across time steps to enforce similarity of suitably-scaled rotation and translation parameters (defined in Section~\ref{sec DM}) as follows:

% The success of traditional bundle adjustment in tasks requiring high accuracy can be attributed to accurate initialization and coherent geometric consistency. However, in case of geometric image distortions, bundle adjustment loses its advantages when relative pose estimation and self-supervised depth are involved in the training process. For example, strict geometric constraints (\textit{e.g.}, epipolar geometry or the fundamental matrix), must be precisely satisfied by the model. Any distortion, noise, or inconsistent depth information could result in poor convergence during training. However, flexible bundle adjustment is designed specifically for this scenario. We propose a regularization term across time steps to enforce similarity of suitably-scaled rotation and translation parameters (defined in Section~\ref{sec DM}) as follows:
%camera smoothness regularization term combined with $\mathbf{\Phi}_i$ and $\mathbf{t}_i$ during the flexible training process:
% \jlk{Are the cameras ordered? Only similarity between neighbors makes sense.}
%
\begin{equation}
  \mathcal{R}_{S}(\pi_{i}) = \frac{1}{N}\sum^{N}_{i=1}(\sum_{j=1}^{T}||\mathbf{\Phi}^{j+1}_{i} - \mathbf{\Phi}^{j}_{i}||^2 + \sum_{j=1}^{T}||\mathbf{t}^{j+1}_{i} - \mathbf{t}^{j}_{i}||^2)
\end{equation}

\noindent where $T$ is the number of time steps for each training batch. Incorporating a pose-update similarity regularization term fosters smooth transitions between adjacent frames, mitigating abrupt variations and ensuring consistency. This regularization enhances optimization stability by constraining pose changes, minimizing oscillations and divergence. Additionally, it aids in mitigating error propagation by safeguarding against significant estimation errors in one frame affecting subsequent frames. 

% \noindent Furthermore, in order to avoid overfitting of ($k_1,k_2,k_3$) parameters, we introduce $L_2$ regularization:
% %
% \begin{equation}
%   \mathcal{R}_{L_2}(\pi_i) = \sum_{i=1}^T||k_{1}||^2 + \sum_{i=1}^T||k_{2}||^2 + \sum_{i=1}^T||k_{3}||^2
% \end{equation}

% \jlk{Summations are over $i$ but parameters do not depend on $i$. See my comment about $\pi_i$ after eq. (8).}

\subsubsection{Fisheye Model Final Loss} We combine the photometric cost mapped into feature space (Eq. \ref{eq fba}) and the camera regularization terms for flexible bundle adjustment training with distorted camera poses:
% \jlk{I think talking of training FBA makes no sense. FBA has no parameters, right? We are training to obtain NeRF parameters and all camera parameters $\pi_i$, right? We are not really training FBA. Am I missing something?}
%
%\vspace{10pt}
\begin{equation}
  \mathcal{L}_{FBA} =  C(x) + \lambda \mathcal{R}_{S}(\pi_i)
  \label{eq fbaf}
\end{equation}
%
%\vspace{10pt}
where $\lambda$ = 0.5. This FBA loss function is only used for fisheye views. 

\subsubsection{Fisheye Depth} We experimented with fisheye depth estimation, but encountered challenges with convergence during training, leading to disappointing results. Fisheye images often have significant distortions and wide-angle perspectives, making accurate depth estimation more difficult compared to pinhole cameras. However, despite the absence of depth information, the final rendering results for fisheye novel views show promise. 

\clearpage

\subsection{Training Objectives}

In supervised training with ground-truth depth, we use the metric-bin module \cite{bhat2023zoedepth} and SI log-loss to improve depth quality. In self-supervised depth optimization, we use an edge-adaptive smoothness loss \cite{chen2023dbarf} {\it via} pretrained depth priors (for pinhole images only): 
\begin{equation}
    \mathcal{L}_{depth} = |\partial x D|e^{-|\partial x \mathcal{I}|} + |\partial x D|e^{-|\partial x \mathcal{I}|}
\end{equation}

\noindent In self-supervised training for camera pose refinement, we use the following photometric loss \cite{gu2023dro}:
\begin{equation}
    %\mathcal{L}_{photo} = \frac{1}{N}\sum_{i=1}^{N}\alpha \frac{1-SSIM(\mathcal{I}^{'}_{i},\mathcal{I}_{i})}{2}+(1-\alpha)||\mathcal{I}^{'}_{i}-\mathcal{I}_{i}||
    \mathcal{L}_{photo} = \frac{1}{N}\sum_{j=1}^{N}\alpha \frac{1-SSIM(\mathcal{I}'_{j},\mathcal{I}_{o})}{2}+(1-\alpha)||\mathcal{I}'_{j}-\mathcal{I}_{o}||_2^2
\label{eq photometric}
\end{equation}
%
% \jlk{Summation over $i=1,...,N$?}
where SSIM is the structural similarity index \cite{1284395} and $\mathcal{I}'_{j}$ is the $j$-th neighboring view warped to the target view $\mathcal{I}_o$. As the photometric error between the target image rendered by IBRNet from nearby views, namely $\widehat{\mathcal{I}}_{i}$, and the ground-truth target image $\mathcal{I}_i$, we use \cite{wang2021ibrnet}:
\begin{equation}
  %\mathcal{L}_{rgb} = \sum^{N}_{i}\sum_{x}||\mathcal{I}^{'}_{i}-\mathcal{I}_{i}(x)||
  \mathcal{L}_{rgb} = ||\widehat{\mathcal{I}}_{i}-\mathcal{I}_{i}||_2^2
\end{equation}
The final pinhole-image loss function to optimize is:
\begin{equation}
    \mathcal{L}_{pinhole} = e^{\beta \cdot t}(\mathcal{L}_{depth}+\mathcal{L}_{photo}) + (1-e^{\beta \cdot t})\mathcal{L}_{rgb}
\label{eq final}
\end{equation}
%
% \jlk{This is final loss for pinhole. What is it for fisheye?}
where $\beta$ = -1$e$4 and $t$ is the current training iteration number. For fisheye images, we use flexible bundle adjustment loss $\mathcal{L}_{FBA}$ (Eq. \ref{eq fbaf}) to achieve faster convergence and ensure consistency. We optimize the  following fisheye-image loss function with $\beta$ = -1$e$3:
\begin{equation}
    \mathcal{L}_{fisheye} = e^{\beta \cdot t}(\mathcal{L}_{FBA}+\mathcal{L}_{photo}) + (1-e^{\beta \cdot t})\mathcal{L}_{rgb}
\label{eq final1}
\end{equation}

\section{Experiments}

\subsubsection{Training} We train Zoe-NK \cite{bhat2023zoedepth} on NYU-depth-v2 dataset to enforce depth priors during planar-pose optimization. We pretrain IBRNet \cite{wang2021ibrnet}, DBARF \cite{chen2023dbarf} and our method on 70\% of LLFF \cite{mildenhall2019local} and 80\% of NeRF\_360\_v2 \cite{barron2022mipnerf}. Only IBRNet uses pseudo-absolute ground-truth camera poses from COLMAP. We also train SCNeRF and our method on a natural fisheye NeRF dataset \cite{jeong2021selfcalibrating} and SCNeRF, OMNI-NeRF, and our method on 70\% of a synthetic fisheye NeRF dataset \cite{eichenseer2022data}.

\subsubsection{Evaluation} We evaluate IBRNet, DBARF, and our method on the LLFF (forward facing) \cite{mildenhall2019local} and NeRF\_360\_v2 \cite{barron2022mipnerf} pinhole-image datasets. On the other hand, we evaluate SCNeRF \cite{jeong2021selfcalibrating}, OMNI-NeRF \cite{shen2022omninerf} and our method on fisheye NeRF datasets \cite{eichenseer2022data, jeong2021selfcalibrating}. In the supplementary material, we also provide results on a self-collected dataset to demonstrate FBINeRF's generalization ability. We perform testing of the pinhole model on 10\% and 5\% of images from LLFF and NeRF\_360\_v2, respectively (not used during pre-training). Other images are reserved for per-scene fine-tuning. We evaluate SCNeRF \cite{jeong2021selfcalibrating} and our method using 5-fold cross-validation on the natural fisheye NeRF dataset, and we evaluate SCNeRF, OMNI-NeRF and our method on 20\% of the synthetic fisheye NeRF dataset. Additional results are available in the supplementary material.

\subsection{Novel-View Synthesis -- Pinhole}

\subsubsection{Depth Priors} 
\begin{figure}[h]
\centering
\includegraphics[scale=0.35]{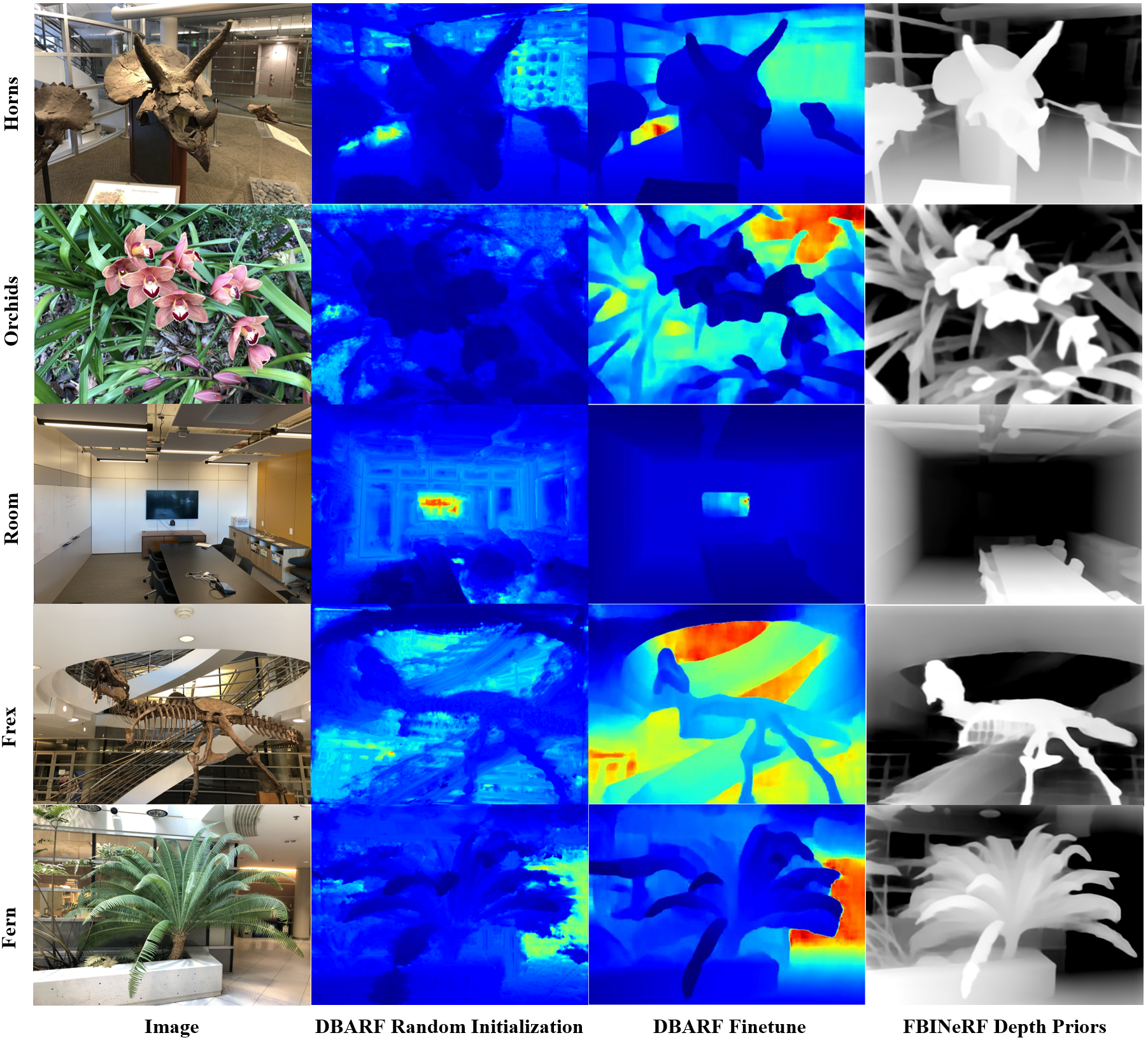} 
\caption{\textbf{Qualitative comparison of depth maps predicted from LLFF dataset} \cite{mildenhall2019local}. FBINeRF depth maps (produced by depth priors and adaptive GRU optimizer) are more accurate than from DBARF (generated by depth header) even after finetune.}
%\vspace{-15pt} 
% \caption{An overview of relative and metric depth combined with flexible bundle adjusting to render novel synthetic views. RGB images are input into the MiDaS depth estimation framework\cite{bhat2023zoedepth} to obtain both relative and metric depth information. Initial pose estimates and sparse 3D points are derived using COLMAP. Subsequently, the refined camera poses and depth maps from flexible bundle adjusting are utilized to project 3D points onto 2D images. The culmination of these steps leads to the rendering of the final NeRF scenes.}
\label{fig 3}
\centering
\end{figure}

Fig.~\ref{fig 3} shows depth maps predicted by DBARF's depth optimizer and by FBINeRF with ZoeDepth-like pretrained model. The results demonstrate our method's ability to recover more accurate depth compared to DBARF (a quantitative comparison is not possible since no ground-truth depth maps are available in LLFF). 

\noindent Moreover, adding a ZoeDepth-like model with metric-bins module extends the model allowing it to learn (including fine-tuning) from ground-truth depth, if available. This applies to both supervised and self-supervised scenarios in FBINeRF, while DBARF is designed for self-supervised depth scenarios only. Next we analyze both qualitative and quantitative results of the pinhole camera model.

% \begin{figure}[!htb]
% \centering
% %\captionsetup{skip=5pt}
% \includegraphics[scale=0.35]{360_result.png} 
% \vspace{-8pt} 
% \caption{\textbf{Qualitative comparison of novel views generated from NeRF\_360\_v2 dataset} \cite{barron2022mipnerf}. Both IBRNet and DBARF produce evident distortions on the bicycle wheels (top), and on the table top and chair (bottom) unlike FBINeRF.}
% \label{fig 4}
% \centering
% \end{figure}

% \vspace{-15pt}

\subsubsection{Ablation Tables} We use PSNR, SSIM \cite{1284395}, and LPIPS \cite{zhang2018unreasonable} as the metrics for novel-view quality assessment. Table \ref{tab 1} shows these metrics for DBARF, IBRNet and our method on the LLFF dataset.

% \vspace{-15pt}

\begin{table}[h]
\centering
\small
\renewcommand{\arraystretch}{1} % 
\caption{\textbf{Quantitative comparison of novel-view synthesis for LLFF dataset} \cite{mildenhall2019local}. Each column shows results either with (\textbf{w}) or without (\textbf{w/o}) per-scene fine-tuning. IBRNet \cite{wang2021ibrnet} and DBARF \cite{chen2023dbarf} results are from the corresponding original papers.}
\scalebox{0.68}{%
\begin{tabular}{lccccccccccccccccccc}
\Xhline{2\arrayrulewidth}
\multirow{3}{*}{Scenes} & \multicolumn{6}{c}{\textbf{PSNR{\(\uparrow\)}}} & \multicolumn{6}{c}{\textbf{SSIM{\(\uparrow\)}}} & \multicolumn{6}{c}{\textbf{LPIPS{\(\downarrow\)}}}\\
\cmidrule(lr){2-7} \cmidrule(lr){8-13} \cmidrule(lr){14-19}
&\multicolumn{2}{c}{IBRNet} &\multicolumn{2}{c}{DBARF} &\multicolumn{2}{c}{Ours} &\multicolumn{2}{c}{IBRNet} &\multicolumn{2}{c}{DBARF} &\multicolumn{2}{c}{Ours}&\multicolumn{2}{c}{IBRNet} &\multicolumn{2}{c}{DBARF} &\multicolumn{2}{c}{Ours}\\
&w/o&w&w/o&w&w/o&w&w/o&w&w/o&w&w/o&w&w/o&w&w/o&w&w/o&w \\
\Xhline{1\arrayrulewidth}  
Fern    &\textbf{23.61}&25.56  &23.12 &25.97 &23.08&\textbf{27.91}  &\textbf{0.743} &0.825&0.724 &0.840&0.733 &\textbf{0.842} &\textbf{0.240}&0.139&0.277&0.120&0.279&\textbf{0.117}\\
Flower  &22.92&23.94  &21.89 &\textbf{23.95} &\textbf{22.97}&23.90  &\textbf{0.849} &0.895&0.793 &\textbf{0.895}&0.812 &0.889 &\textbf{0.123}&0.074&0.176&0.074&0.148&\textbf{0.071}\\
Fortress&\textbf{29.05}&31.18  &28.13 &31.43 &28.88&\textbf{32.58}  &0.850 &0.918&0.820 &\textbf{0.918}&\textbf{0.863} &0.917 &\textbf{0.087}&0.046&0.126&0.046&0.098&\textbf{0.042}\\
Horns   &24.96&\textbf{28.46}  &24.17 &27.51 &\textbf{25.01}&26.33  &\textbf{0.831} &\textbf{0.913}&0.799 &0.903&0.801 &0.903 &0.144&\textbf{0.070}&0.194&0.076&\textbf{0.121}&0.073\\
Leaves  &\textbf{19.03}&\textbf{21.28}  &18.85 &20.32 &18.70&19.98  &0.737 &\textbf{0.807}&0.649 &0.758&\textbf{0.752} &0.744 &0.289&\textbf{0.137}&0.313&0.156&\textbf{0.271}&0.147\\
Orchids &\textbf{18.52}&20.83  &17.78 &20.26 &18.33&\textbf{21.92}  &\textbf{0.573} &\textbf{0.722}&0.506 &0.693&0.529 &0.691 &\textbf{0.259}&0.142&0.352&0.151&0.357&\textbf{0.138}\\
Room    &28.81&31.05  &27.50 &31.09 &\textbf{28.90}&\textbf{32.21}  &0.926 &\textbf{0.950}&0.901 &0.947&\textbf{0.934} &0.949 &0.099&\textbf{0.060}&0.142&0.063&\textbf{0.089}&0.061\\
Trex    &\textbf{23.51}&\textbf{26.52}  &22.70 &22.82 &22.03&24.77  &\textbf{0.818} &0.905&0.783 &0.848&0.794 &\textbf{0.912} &\textbf{0.160}&\textbf{0.074}&0.207&0.120&0.168&0.084\\
\Xhline{1\arrayrulewidth}
Average   &\textbf{23.80}&26.10&23.02&25.42&23.49&\textbf{26.20}&\textbf{0.791}&\textbf{0.867}&0.747&0.850&0.777&0.856&\textbf{0.175}&0.093&0.223&0.101&0.191&\textbf{0.092}\\
\Xhline{1\arrayrulewidth} 
\end{tabular}%
}
\label{tab 1}
% \vspace{-10pt} 
\end{table}

\vspace{-25pt}

\subsubsection{Analysis} Table \ref{tab 1} shows these metrics for DBARF, IBRNet and our method on the LLFF dataset. Our method outperforms DBARF in the {\it average} value of each metric both with and without per-scene fine-tuning. While without per-scene fine-tuning our method is outperformed by IBRNet, this is not surprising since IBRNet is tuned to the LLFF dataset and uses absolute camera poses whereas our method {\it estimates} relative camera poses.
% However, with per-scene fine-tuning our method outperforms IBRNet in terms of average PSNR and LPIPS, despite not using absolute camera poses.

% \vspace{-15pt}

\begin{table}[h]
\centering
\small
\renewcommand{\arraystretch}{1} % 
\caption{\textbf{Training time comparison of novel-view synthesis for LLFF dataset} \cite{mildenhall2019local}. We compare time consumption in view selection and pose refinement between DBARF and our method. IBRNet uses absolute pose and no depth is involved, so we exclude it.}
\scalebox{0.9}{%
\begin{tabular}{lccccccccccccccccccc}
\Xhline{2\arrayrulewidth}
\multirow{3}{*}{Scenes} & \multicolumn{4}{c}{\textbf{View Select}} & \multicolumn{4}{c}{\textbf{Feature Extract}} & \multicolumn{4}{c}{\textbf{Pose Refine}}\\
\cmidrule(lr){2-5} \cmidrule(lr){6-9} \cmidrule(lr){10-13}
&\multicolumn{2}{c}{DBARF} &\multicolumn{2}{c}{Ours} &\multicolumn{2}{c}{DBARF} &\multicolumn{2}{c}{Ours} &\multicolumn{2}{c}{DBARF} &\multicolumn{2}{c}{Ours}\\
&w/o&w&w/o&w&w/o&w&w/o&w&w/o&w&w/o&w\\
\Xhline{1\arrayrulewidth}  
LLFF    &4.62 &4.78 &\textbf{1.33} &1.65 &\textbf{1.94} &1.98 &3.56 &3.75 &5.12 &5.73 &\textbf{2.68} &2.89\\
\Xhline{1\arrayrulewidth}
% Average   &\textbf{23.80}&26.10&23.02&25.42&23.49&\textbf{26.20}&\textbf{0.791}&\textbf{0.867}\\
% \Xhline{1\arrayrulewidth} 
\end{tabular}%
}
\label{tab 2}
%\vspace{-10pt}
\end{table}

\vspace{-25pt}

\subsubsection{Analysis} Table~\ref{tab 2} reveals that DBARF requires more time for view selection and pose refinement compared to our method. However, it excels in feature extraction since it uses a ResNet-like backbone. The values in Table~\ref{tab 2} represent a relative time cost of training. Both methods processed the entire LLFF dataset and generated rendered views for all scenes. Overall, the training time required by our method is notably smaller than that of DBARF.

% Fern    &\textbf{23.61}&25.56  &23.12 &25.97 &23.08&\textbf{27.91}  &\textbf{0.743} &0.825\\
% Flower  &22.92&23.94  &21.89 &\textbf{23.95} &\textbf{22.97}&23.90  &\textbf{0.849} &0.895\\
% Fortress&\textbf{29.05}&31.18  &28.13 &31.43 &28.88&\textbf{32.58}  &0.850 &0.918\\
% Horns   &24.96&\textbf{28.46}  &24.17 &27.51 &\textbf{25.01}&26.33  &\textbf{0.831} &\textbf{0.913}\\
% Leaves  &\textbf{19.03}&\textbf{21.28}  &18.85 &20.32 &18.70&19.98  &0.737 &\textbf{0.807}\\
% Orchids &\textbf{18.52}&20.83  &17.78 &20.26 &18.33&\textbf{21.92}  &\textbf{0.573} &\textbf{0.722}\\
% Room    &28.81&31.05  &27.50 &31.09 &\textbf{28.90}&\textbf{32.21}  &0.926 &\textbf{0.950}\\
% Trex    &\textbf{23.51}&\textbf{26.52}  &22.70 &22.82 &22.03&24.77  &\textbf{0.818} &0.905\\

% \vspace{-30pt}

% \subsubsection{Views} Here is the result. Here is the result.Here is the result.Here is the result.Here is the result.Here is the result.Here is the result.Here is the result.Here is the result.Here is the result.Here is the result.Here is the result.Here is the result.Here is the result.Here is the result.

% \begin{figure}[h]
% \centering
% \includegraphics[scale=0.25]{360New.png} 
% \caption{\textbf{Qualitative comparison of novel views generated from NeRF\_360\_v2 dataset} \cite{barron2022mipnerf}. DBARF produce evident distortions on the bicycle wheels (top), and on the table top and chair (bottom) unlike FBINeRF.}
% \label{fig 4}
% \centering
% \end{figure}

\subsection{Novel-View Synthesis -- Fisheye}

\subsubsection{Synthetic Dataset} 
\begin{figure}[h]
\centering
%\captionsetup{skip=5pt}
\includegraphics[scale=0.36]{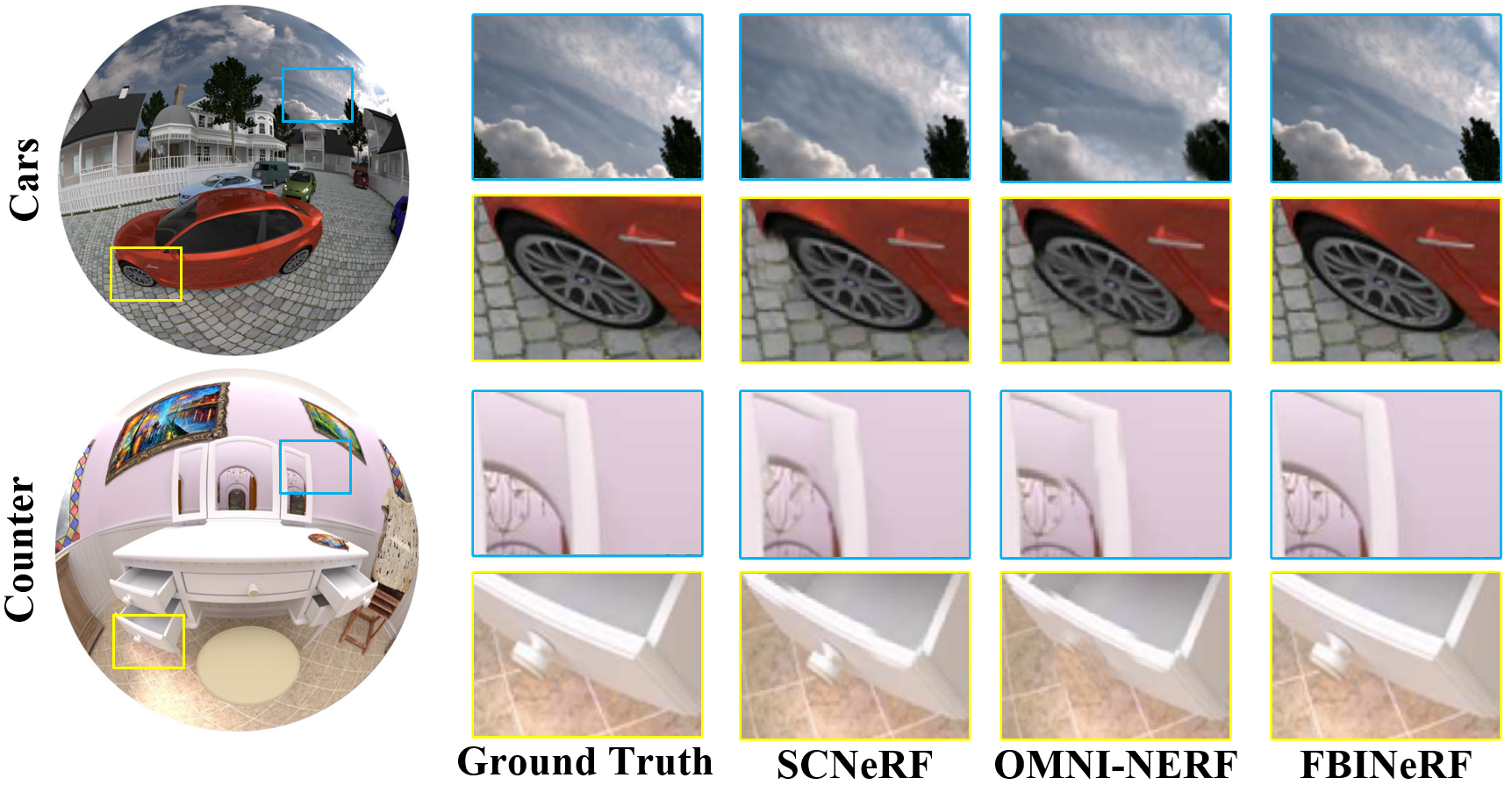}
%\vspace{-8pt} 
\caption{\textbf{Qualitative comparison of novel views generated from a synthetic fisheye NeRF dataset} \cite{eichenseer2022data}. Both SCNeRF and OMNI-NeRF produce clear distorions in the clouds, tree and car wheel (top) and on the mirror and drawer (bottom) whereas FBINeRF does not.}
\label{fig 4}
\centering
\end{figure}

Fig.~\ref{fig 4} demonstrates improvements offered by our method over SCNeRF and OMNI-NeRF on scenes from a synthetic fisheye NeRF dataset \cite{eichenseer2022data}. While both SCNeRF and OMNI-NeRF produce very visible distorions in the clouds, tree and car wheel (top images) and on the mirror and drawer (bottom images), FBINeRF generates very accurate renderings. 

% \begin{figure}[h]
% \centering
% %\captionsetup{skip=5pt}
% \includegraphics[scale=0.3]{synthesis_fisheye.png}
% %\vspace{-8pt} 
% \caption{\textbf{Qualitative comparison of novel views generated from a synthetic fisheye NeRF dataset} \cite{eichenseer2022data}. Both SCNeRF and OMNI-NeRF produce clear distorions in the clouds, tree and car wheel (top) and on the mirror and drawer (bottom) whereas FBINeRF does not.}
% \label{fig 4}
% \centering
% \end{figure}
\vspace{-30pt}

\begin{table}[!h]
\begin{center}
\small
\renewcommand{\arraystretch}{1.2} % 
\caption{\textbf{Quantitative comparison of novel-view synthesis for a natural fisheye NeRF dataset} \cite{jeong2021selfcalibrating}. The FBINeRF results here are for spherical ray sampling while the third column of images in Fig.~\ref{fig 5} are rendered via rectified poses.}
\scalebox{0.7}{%
\begin{tabular}{lccccccc}
\Xhline{2\arrayrulewidth}
\multirow{2}{*}{Scenes} & \multicolumn{2}{c}{\textbf{PSNR{\(\uparrow\)}}} & \multicolumn{2}{c}{\textbf{SSIM{\(\uparrow\)}}} & \multicolumn{2}{c}{\textbf{LPIPS{\(\downarrow\)}}}\\
\cmidrule(lr){2-3} \cmidrule(lr){4-5} \cmidrule(lr){6-7}
&\multicolumn{1}{c}{SCNeRF} &\multicolumn{1}{c}{FBINeRF} &\multicolumn{1}{c}{SCNeRF} &\multicolumn{1}{c}{FBINeRF} &\multicolumn{1}{c}{SCNeRF} &\multicolumn{1}{c}{FBINeRF} \\
\Xhline{1\arrayrulewidth}
Chair   & 17.62 & \textbf{24.29} & 0.312        & \textbf{0.428}    &\textbf{0.129}    &0.134\\
Rock    & 18.91 & \textbf{25.18} & 0.201        & \textbf{0.225}    &0.223    &\textbf{0.112}\\
Heart   & 25.33 & \textbf{27.94} & \textbf{0.520} & 0.402             &0.451    &\textbf{0.089}\\
Flowers & \textbf{28.19} &26.73  & 0.792        & \textbf{0.822}    &\textbf{0.197}    &0.378\\
Globe   & 23.70 & \textbf{27.67} & 0.838        & \textbf{0.910}    &0.210    &\textbf{0.096}\\
Cube    & 27.44 & \textbf{29.22} & 0.692        & \textbf{0.793}    &0.514    &\textbf{0.081}\\
\Xhline{2\arrayrulewidth}
Average & 23.53 & \textbf{26.83} & 0.559 & \textbf{0.597} & 0.287 & \textbf{0.148}\\
\Xhline{2\arrayrulewidth}
\end{tabular}
}
%\vspace{-5pt}
\label{tab 3}
\end{center}
%\vspace{-25pt} % Adjust the value to your preference
\end{table}

\vspace{-20pt}

\subsubsection{Fisheye NeRF Dataset} 
\begin{figure}[!h]
\centering
%\captionsetup{skip=5pt}
\includegraphics[scale=0.57]{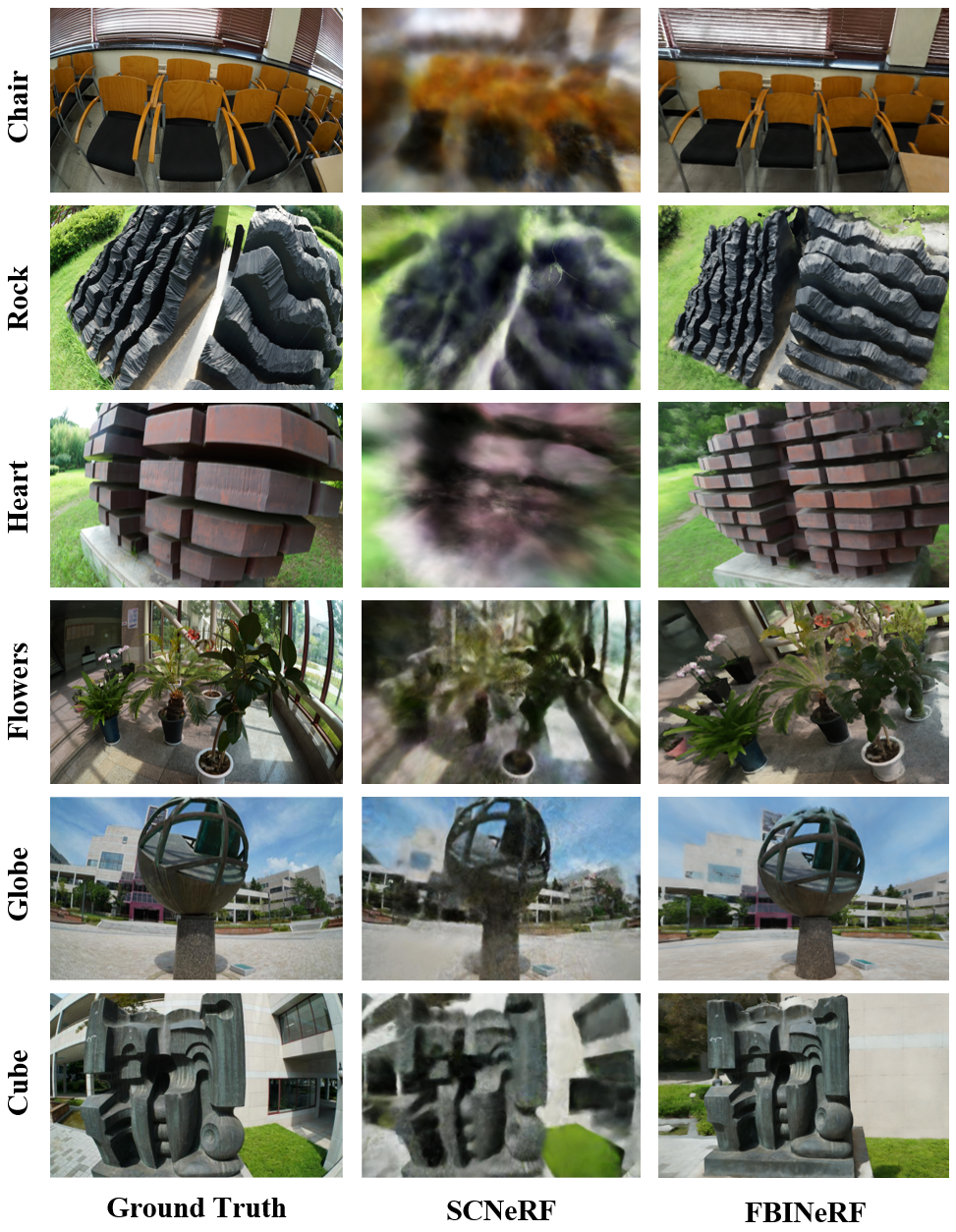}
%\vspace{-8pt} 
\caption{\textbf{Qualitative results on fisheye NeRF dataset}\cite{jeong2021selfcalibrating} for SCNeRF and FBINeRF. We show qualitative results for our method and SCNeRF on the fisheye NeRF dataset \cite{jeong2021selfcalibrating}. As we can see, our method produces sharper, less noisy, and more accurate rendered views than SCNeRF. On average, SCNeRF requires over 6 hours to train for each scene in this dataset to obtain results shown in Fig. \ref{fig 5} while ours generates continuous dense voxel fisheye novel views within half an hour.}
\label{fig 5}
\centering
\end{figure}

Fig.~\ref{fig 5} shows novel-view generation results on a natural fisheye NeRF dataset \cite{jeong2021selfcalibrating} for SCNeRF and our method. Very obvious blurring can be seen in the views rendered by SCNeRF while those generated by FBINeRF are sharp and very close to the ground truth. However, we could not synthesize novel views from this dataset using OMNI-NeRF due to a mismatch between parameters available in the dataset and those required by OMNI-NeRF. We would like to point out that SCNeRF necessitates long training. One possible reason is its use of the projected ray distance loss function which requires volume rendering at each training iteration (each novel fisheye-view scene requires almost 6 hours to be generated without continuous rendering of views). 

%\vspace{-20pt}

\subsubsection{Meshes} 
\begin{figure}[h]
\centering
%\captionsetup{skip=5pt}
\includegraphics[scale=0.35]{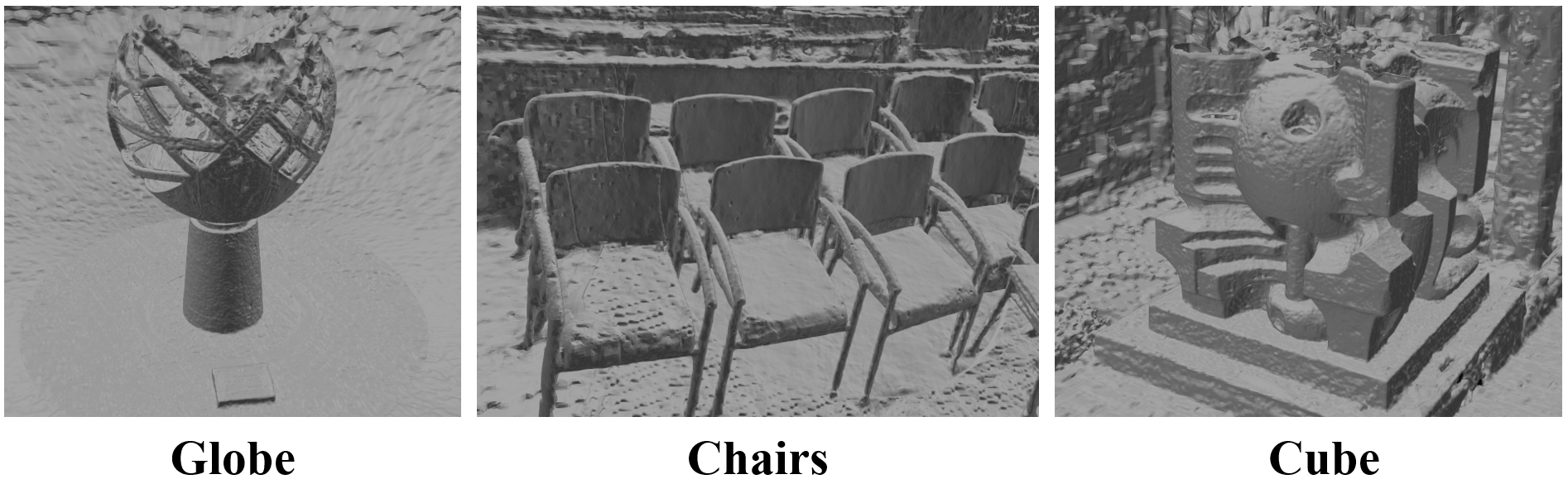}
%\vspace{-8pt} 
\caption{\textbf{Qualitative mesh results from fisheye NeRF dataset}\cite{jeong2021selfcalibrating} produced by FBINeRF. Meshes from FBINeRF can be imported into Unity, Unreal Engine, Blender software for rendering or other downstream tasks.}
\label{fig 6}
\centering
\end{figure}

Figure~\ref{fig 6} shows examples of meshes produced FBINeRF that can be used as a dense representation for downstream tasks.

\section{Limitation and Discussion}

Our method can produce high-fidelity synthesic views for both pinhole and fisheye cameras. However, the presence of abnormal geometric-distortion parameters can lead to failures in the inference stage when using pre-trained models. This is because fisheye images require more precise pose estimation, leading to reduced generalization of our method. In principle, depth maps in a fisheye-image scenario could be jointly optimized with camera poses. However, a simple third-order polynomial approximation of geometric distortions cannot accurately model images from a fisheye camera. In our future work, we are planing to research self-supervised depth methods for fisheye cameras that can be incorporated into our method to speed up the training process. We are also planning to incorporate more accurate lens-distortion modeling.

\section{Conclusion}
In this work, we have identified persistent challenges that SOTA NeRF methods face. While generalized NeRF methods are sensitive to inaccurate depth initialization and are inefficient in updating feature cost maps, fisheye NeRF methods suffer from severe visual artifacts due to radial distortions in the input views. To address these issues, we developed FBINeRF that incorporates DenseNet and AdamR-GRU-like optimizers to enhance pose optimization. FBINeRF also improves depth initialization by introducing depth priors through a ZoeDepth-like pretrained model thus facilitating both supervised and self-supervised scenarios. Preliminary results of our attempt to integrate a feature-based deep recurrent neural network with flexible-bundle adjustment for fisheye-image modeling in NeRF are also promising, suggesting the potential for further enhancements. 

% \clearpage\mbox{}Page \thepage\ of the manuscript.
% \clearpage\mbox{}Page \thepage\ of the manuscript.
% \clearpage\mbox{}Page \thepage\ of the manuscript.
% \clearpage\mbox{}Page \thepage\ of the manuscript.
% \clearpage\mbox{}Page \thepage\ of the manuscript. This is the last page.
% Now we have reached the maximum length of an ECCV \ECCVyear{} submission (excluding references).
% References should start immediately after the main text, but can continue past p.\ 14 if needed.
\clearpage  % TODO REVIEW/FINAL: This \clearpage needs to be removed from both review and camera-ready versions.

\bibliographystyle{splncs04}
\bibliography{ref}
\end{document}